\documentclass{article}

\PassOptionsToPackage{numbers, compress}{natbib}

\usepackage[final]{neurips_2018}

\usepackage[utf8]{inputenc} 
\usepackage[T1]{fontenc}    
\usepackage{hyperref}       
\usepackage{url}            
\usepackage{booktabs}       
\usepackage{amsfonts}       
\usepackage{nicefrac}       
\usepackage{microtype}      
\usepackage{subcaption} 
\usepackage{cleveref}
\usepackage{amssymb}
\usepackage{amsmath}
\usepackage{bm}
\usepackage{wrapfig}
\usepackage{authblk}
\usepackage[toc,page]{appendix}

\newcommand{\module}[1]{{\color{nice-purple!75!black}\verb~#1~}}

\renewcommand\vec[1]{{\bm{#1}}}

\usepackage{xcolor}
\definecolor{nice-red}{HTML}{E41A1C}
\colorlet{dark-red}{nice-red!80!black}
\definecolor{nice-orange}{HTML}{FF7F00}
\colorlet{dark-orange}{orange!85!black}
\definecolor{nice-yellow}{HTML}{FFC020}
\definecolor{nice-green}{HTML}{4DAF4A}
\definecolor{nice-blue}{HTML}{377EB8}
\definecolor{nice-purple}{HTML}{984EA3}

\usepackage{tikz}
\usetikzlibrary{calc,trees,positioning,arrows,chains,shapes.geometric,%
  decorations.pathreplacing,decorations.pathmorphing,shapes,%
  matrix,shapes.symbols,fit,decorations,arrows.meta}

\usepackage{adjustbox}
\usepackage{color, soul}

\definecolor{pink}{RGB}{226, 113, 187}

\newcommand{\eInferSent}{e-\textsc{InferSent}}
\newcommand{\inferSentAutoencoder}{\textsc{InferSentAutoEnc}}
\newcommand{\expzero}{\textsc{PremiseAgnostic}}
\newcommand{\expone}{\textsc{PredictAndExplain}}
\newcommand{\exptwo}{\textsc{ExplainThenPredict}}
\newcommand{\exptwoseqtoseq}{\textsc{ExplainThenPredictSeq2Seq}}
\newcommand{\exptwoattention}{\textsc{ExplainThenPredictAttention}}
\newcommand{\expthree}{\textsc{Represent}}
\newcommand{\expfour}{\textsc{Transfer}}

\title{e-SNLI: Natural Language Inference with \\ Natural Language Explanations}

\author{
  \textbf{Oana-Maria Camburu}\textsuperscript{1         } 
   \textbf{Tim Rockt\"aschel}\textsuperscript{2         }  
  \textbf{Thomas Lukasiewicz}\textsuperscript{1,3  } 
  \textbf{Phil Blunsom}\textsuperscript{1,4} \\
  \texttt{\{oana-maria.camburu, thomas.lukasiewicz, phil.blunsom\}@cs.ox.ac.uk} \\
  \texttt{t.rocktaschel@ucl.ac.uk} \\
  \textsuperscript{1}Department of Computer Science, University of Oxford\\
  \textsuperscript{2}Department of Computer Science, University College London \\
  \textsuperscript{3}Alan Turing Institute, London, UK \\
  \textsuperscript{4}DeepMind, London, UK
  
}

\begin{document}

\maketitle

\begin{abstract}
  In order for machine learning to garner widespread public adoption, models must be able to provide interpretable and robust explanations for their decisions, as well as learn from human-provided explanations at train time. In this work, we extend the Stanford Natural Language Inference dataset with an additional layer of human-annotated natural language explanations of the entailment relations. We further implement models that incorporate these explanations into their training process and output them at test time. We show how our corpus of explanations, which we call e-SNLI, can be used for various goals, such as obtaining full sentence justifications of a model's decisions, improving universal sentence representations and transferring to out-of-domain NLI datasets. Our dataset\footnote{https://github.com/OanaMariaCamburu/e-SNLI} thus opens up a range of research directions for using natural language explanations, both for improving models and for asserting their trust.
\end{abstract}
\section{Introduction}
Humans do not learn solely from labeled examples supplied by a teacher.
Instead, they seek a conceptual understanding of a task through both demonstrations and explanations.
Machine learning models trained simply to obtain high accuracy on held-out sets often learn to rely heavily on shallow input statistics, resulting in brittle models susceptible to adversarial attacks. For example, \citet{why-trust-you-RibeiroSG16} present a document classifier that distinguishes between {\it Christianity} and {\it Atheism} with an accuracy of $94\%$. However, on close inspection, the model spuriously separates classes based on words contained in the headers, such as \textit{Posting}, \textit{Host}, and \textit{Re}. 

In this work, we introduce a new dataset and models for exploiting and generating explanations for the task of recognizing textual entailment. We argue for free-form natural language explanations, as opposed to formal language, for a series of reasons. First, natural language is readily comprehensible to an end-user who needs to assert a model's reliability. Secondly, it is also easiest for humans to provide free-form language, eliminating the additional effort of learning to produce formal language, thus making it simpler to collect such datasets. 
Lastly, natural language justifications might eventually be mined from existing large-scale free-form text. 

Despite the potential for free-form justifications to improve both learning and transparency, there is currently a lack of such datasets in the machine learning community. 
To address this deficiency, we have collected a large corpus of human-annotated explanations for the Stanford Natural Language Inference (SNLI) dataset~\cite{snli}. We chose SNLI because it constitutes an influential corpus for natural language understanding that requires deep assimilation of fine-grained nuances of common-sense knowledge. 
We call our explanation-augmented dataset e-SNLI, which we collected to enable research in the direction of training with and generation of free-form textual justifications. 

In order to demonstrate the efficacy of the e-SNLI dataset, we first show that it is much more difficult to produce correct explanations based on spurious correlations than to produce correct labels. We then implement models that, given a premise and a hypothesis, predict a label and an explanation. We also investigate how the additional signal from explanations received at train time can guide models into learning better sentence representations. Finally, we look into the transfer capabilities of our model to out-of-domain NLI datasets.

\section{Background}

The task of recognizing textual entailment is a critical natural language understanding task. Given a pair of sentences, called the premise and hypothesis, the task consists of classifying their relation as either (a)~{\it entailment}, if the premise entails the hypothesis, (b)~{\it contradiction}, if the hypothesis contradicts the premise, or (c)~{\it neutral}, if neither entailment nor contradiction hold. The SNLI dataset \cite{snli}, containing $570K$ data points of human-generated triples (premise, hypothesis, label), has driven the development of a large number of neural network models \cite{RocktaschelGHKB15, snli-1, snli-2, esim, snli-4, kim, infersent}. 

\citet{infersent} showed that training universal sentence representations on SNLI is both more efficient and more accurate than the traditional training approaches on orders of magnitude larger, but unsupervised, datasets \cite{skip, fastsent}. 
We take this approach one step further and show that an additional layer of explanations on top of the label supervision brings further improvement. 

Recently, \citet{artifacts} cast doubt on whether models trained on SNLI are learning to understand language, or are largely fixating on spurious correlations, also called artifacts. For example, specific words in the hypothesis tend to be strong indicators of the label, e.g., {\it friends, old} appear very often in neutral hypotheses, {\it animal, outdoors} appear most of the time in entailment hypotheses, while {\it nobody, sleeping} appear mostly in contradiction hypothesis. 
They show that a premise-agnostic model, i.e., a model that only takes as input the hypothesis and outputs the label, obtains $67\%$ test accuracy. In section \ref{expzero} we show that it is much more difficult to rely on artifacts to generate explanations than to generate labels.  

\section{Collecting explanations}
\begin{figure}
  \centering
  \begin{tabular}{l}
    \toprule

    Premise: An adult dressed in black \hl{holds a stick}. \\ 
    Hypothesis: An adult is walking away, \hl{empty-handed}. \\ 
    Label: contradiction\\ 
    Explanation: Holds a stick implies using hands so it is not empty-handed.\\

    \midrule
    
    Premise: A child in a yellow plastic safety swing is laughing as a dark-haired woman \\
    in pink and coral pants stands behind her.\\ Hypothesis:	A young \hl{mother} is playing with her \hl{daughter} in a swing.\\ 
    Label: neutral\\ 
    Explanation: Child does not imply daughter and woman does not imply mother.\\

    \midrule
    
    Premise: A \hl{man} in an orange vest \hl{leans over a pickup truck}. \\ 
    Hypothesis: A man is \hl{touching} a truck.\\ 
    Label: entailment\\ 
    Explanation: Man leans over a pickup truck implies that he is touching it. \\

    \bottomrule
\end{tabular}
  \caption{Examples from e-SNLI. Annotators were given the premise, hypothesis, and label. They highlighted the words that they considered essential for the label and provided the explanations.
}
  \label{examples_table}
\end{figure}

We present our collection methodology for e-SNLI, for which we used Amazon Mechanical Turk. The main question that we want our dataset to answer is: {\it Why is a pair of sentences in a relation of entailment, neutrality, or contradiction?} We encouraged the annotators to focus on the non-obvious elements that induce the given relation,  and not on the parts of the premise that are repeated identically in the hypothesis. For entailment, we required justifications of all the parts of the hypothesis that do not appear in the premise. For neutral and contradictory pairs, while we encouraged stating all the elements that contribute to the relation, we consider an explanation correct, if at least one element is stated. Finally, we asked the annotators to provide self-contained explanations, as opposed to sentences that would make sense only after reading the premise and hypothesis. For example, we prefer an explanation of the form ``\textit{Anyone can knit, not just women.}'', rather than ``\textit{It cannot be inferred they are women.}''

In crowd-sourcing, it is difficult to control the quality of free-form annotations. Thus, we aimed to preemptively block the submission of obviously incorrect answers. We did in-browser checks to ensure that each explanation contained at least three tokens and that it was not a copy of the premise or hypothesis. We further guided the annotators to provide adequate answers by asking them to proceed in two steps. First, we require them to highlight words from the premise and/or hypothesis that they consider essential for the given relation. Secondly, annotators had to formulate the explanation using the words that they highlighted. However, using exact spelling might push annotators to formulate grammatically incorrect sentences, therefore we only required half of the highlighted words to be used with the same spelling. 
For entailment pairs, we required at least one word in the premise to be highlighted. 
For contradiction pairs, we required highlighting at least one word in both the premise and the hypothesis. 
For neutral pairs, we only allowed highlighting words in the hypothesis, in order to strongly emphasize the asymmetry in this relation and to prevent workers from confusing the premise with the hypothesis. 
We believe these label-specific constraints helped in putting the annotator into the correct mindset, and additionally gave us a means to filter incorrect explanations. 
Finally, we also checked that the annotators used other words that were not highlighted, as we believe a correct explanation would need to articulate a link between the keywords.

We collected one explanation for each pair in the training set and three explanations for each pair in the validation and test sets. \Cref{examples_table} shows examples of collected explanations. There were 6325 workers with an average of 860 explanations per worker and a standard deviation of 403.

\textbf{Analysis and refinement of the collected dataset} \label{partial-score} In order to measure the quality of our collected explanations, we selected a random sample of $1000$ examples and manually graded their correctness between $0$ (incorrect) and $1$ (correct), giving partial scores of $k/n$ if only $k$ out of $n$ required arguments were mentioned. We also considered an explanation as incorrect if it was uninformative, that is, if the explanation was template-like, extensively repeating details from the premise/hypothesis that are not directly useful for justifying the relation between the two sentences. We observed a few re-occurring templates such as: ``\textit{Just because [entire premise] doesn't mean [entire hypothesis]}'' for neutral pairs, ``\textit{[entire premise] implies [entire hypothesis]}'' for entailment pairs, and ``\textit{It can either be [entire premise] or [entire hypothesis]}'' for contradiction pairs. We assembled a list of templates, which can be found in Appendix \ref{expl_templates}, that we used for filtering the dataset of such uninformative explanations. Specifically, we filtered an explanation if its edit distance to one of the templates was less than $10$. We ran this template detection on the entire dataset and reannotated the detected explanations ($11\%$ in total).  

Our final counts show a total error rate of $9.62\%$, with $19.55\%$ on entailment, $7.26\%$ on neutral, and $9.38\%$ on contradiction. We notice that entailment pairs were by far the most difficult to obtain proper explanations for. This is firstly due to partial explanations, as annotators had an incentive to provide shorter inputs, so they often only mentioned one argument. A second reason is that many of the entailment pairs have the hypothesis as almost a subset of the premise, prompting the annotators to just repeat that as a statement.

\section{Experiments}
We first present an experiment which demonstrates that a model which can easily rely on artifacts in SNLI to provide correct labels would not be able to provide correct explanations as easily. We refer to it as \expzero.

We then present a series of experiments to elucidate whether models trained on e-SNLI are able to: (i)~predict a label and generate an explanation for the predicted label (referred to as \expone), (ii)~generate an explanation then predict the label given only the generated explanation (\exptwo), (iii)~learn better universal sentence representations (\expthree), and  (iv)~transfer to out-of-domain NLI datasets (\expfour).

Throughout our experiments, our models follow the architecture presented in \citet{infersent}, as we build directly on top of their code\footnote{https://github.com/facebookresearch/InferSent. We fixed the issue raised in https://github.com/ facebookresearch/InferSent/issues/51 that the max-pooling was taken over paddings.}. Therefore, our encoders are 2048-bidirectional-LSTMs \cite{lstm} with max-pooling, resulting in a sentence representation dimension of 4096. Our label classifiers are 3-layers MLPs with 512 internal size and without non-linearities. For our explanation decoders, we used a simple one-layer LSTM, for which we tried internal sizes of $512$, $1024$, $2048$, and $4096$.
In order to reduce the vocabulary size for explanation generation, we replaced words that appeared less than $15$ times\footnote{Counted among premises, hypothesis, and explanations.} with {\tt <UNK>.}
We obtain an output vocabulary of approximately $12K$ words. 
The preprocessing and optimization were kept the same as in \cite{infersent}. 

Whenever appropriate, we run our models with five seeds and provide the average performance with the standard deviation in parenthesis. If no standard deviation is reported, the results are from one experiment with seed 1234. 

\subsection{\expzero: Generate an explanation given only the hypothesis}
\label{expzero}

\citet{artifacts} show that a neural network that only has access to the hypothesis can predict the correct label $67\%$ of the times. We are therefore interested in evaluating how well our explanations can be predicted from hypotheses alone.

\textbf{Model   } We train a 2048-bidirectional-LSTM with max-pooling for encoding the hypothesis, followed by a one-layer LSTM for decoding the explanation. The initial state of the decoder is the vector embedding of the hypothesis, which is also concatenated at every timestep of the decoder, to avoid forgetting.

\textbf{Selection   } We consider internal sizes of the decoder of $512$, $1024$, $2048$ and $4096$. We pick the model that gives the best perplexity on the validation set. We notice that the perplexity strictly decreases when we increase the decoder size. However, for practical reasons, we do not increase the decoder size beyond $4096$.

\textbf{Results   } 
We then manually look at the first $100$ test examples and obtain that only $6.83$\footnote{Partial scoring as explained in Section \ref{partial-score}.} were correct. We also separately train the same hypothesis-only encoder for label prediction alone and obtain $66$ correct labels in the same first $100$ test examples. This validates our intuition that it is much more difficult (approx.\ 10x for this architecture) to rely on spurious correlations to predict correct explanations than to predict correct labels. 


\subsection{\expone: Jointly predict a label and generate an explanation for the predicted label}

In this experiment, we investigate how the typical architecture employed on SNLI can be enhanced with a module that aims to justify the decisions of the entire network.

\textbf{Model   } We employ the InferSent \cite{infersent} architecture, where a bidirectional-LSTM with max-pooling separately encodes the premise, $\vec{u}$, and hypothesis, $\vec{v}$. The vector of features $\vec{f} = [\vec{u}, \vec{v}, |\vec{u} - \vec{v}|, \vec{u} \odot \vec{v}]$ is then passed to the MLP classifier that outputs a distribution over the 3 labels.  We add a one-layer LSTM decoder for explanations, which takes the feature vector $\vec{f}$ both as an initial state and concatenated to the word embedding at each time step.

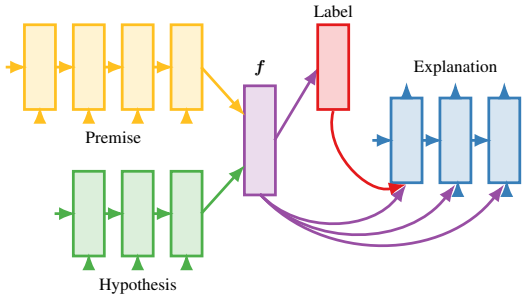
\begin{wrapfigure}{R}{0.5\textwidth}
    \tikzstyle{premise}=[draw, ultra thick, color=nice-yellow, text height=1.5cm, text width=0.4cm, fill=nice-yellow!20]
    \tikzstyle{hypothesis}=[draw, ultra thick, color=nice-green, text height=1.5cm, text width=0.4cm, fill=nice-green!20]
    \tikzstyle{label}=[draw, ultra thick, color=nice-red, text height=1.5cm, text width=0.4cm, fill=nice-red!20]
    \tikzstyle{explanation}=[draw, ultra thick, color=nice-blue, text height=1.5cm, text width=0.4cm, fill=nice-blue!20]
    \tikzstyle{concat}=[draw, ultra thick, color=nice-purple, text height=2cm, text width=0.4cm, fill=nice-purple!20]
    \tikzstyle{connect}=[ultra thick, -Latex]
    
    \centering
    \resizebox{1\linewidth}{!}{
    \begin{tikzpicture}
        \foreach \i in {0,...,3} {
            \node[premise] at (\i, 0) {};
            \draw[connect, nice-yellow] (\i-0.7, 0) -- (\i-0.25, 0);
            \draw[connect, nice-yellow] (\i, -1.1) -- (\i, -0.8);
        }
        \node[anchor=north] at (1.5, -1.2) {Premise};
        
        \begin{scope}[shift={(0,-3)}]
        \foreach \i in {1,...,3} {
            \node[hypothesis] at (\i, 0) {};
            \draw[connect, nice-green] (\i-0.7, 0) -- (\i-0.25, 0);
            \draw[connect, nice-green] (\i, -1.1) -- (\i, -0.8);
        }
        \node[anchor=north] at (2, -1.2) {Hypothesis};
        \end{scope}
        
        \begin{scope}[shift={(4.5,-1.5)}]
        \foreach \i in {0} {
            \node[concat] at (0,\i) {};
        }
        \draw[connect, nice-yellow] (-1.2, 1.5) -- (-0.3, 0.5);
        \draw[connect, nice-green] (-1.2, -1.5) -- (-0.3, -0.5);
        \draw[connect, nice-purple] (0.3, 0) -- (1.2, 1.5);
        \end{scope}
        
        \node[anchor=south, text width=2cm, text centered] at (4.5, -0.3) {$\vec{f}$};
        
        \begin{scope}[shift={(7.5,-1.5)}]
        \foreach \i in {0,...,2} {
            \node[explanation] at (\i, 0) {};
            \draw[connect, nice-blue] (\i-0.7, 0) -- (\i-0.25, 0);
            \draw[connect, nice-blue] (\i, 0.9) -- (\i, 1.2);
            \path[connect, nice-purple] (-3,-1.1) edge[bend right=50] (\i, -0.9);
        }
        \foreach \i in {1,2} {
            \draw[connect, nice-blue] (\i+0.05, -1.1) -- (\i+0.05, -0.8);
        }
        \node[anchor=south] at (1, 1.2) {Explanation};
        \node[label] (label) at (-1.5, 1.5) {}; 
        \path[connect, nice-red] (-1.5,0.6) edge[bend right=60] (0, -0.9);
        \node[above = 0cm of label] {Label};
        \end{scope}
    \end{tikzpicture}
    }
    \caption{Overview of the \eInferSent{} architecture.}
    \label{fig:model}
\end{wrapfigure}

In order to condition the explanation also on the label, we prepend the label as a word (\textit{entailment, contradiction, neutral}) at the beginning of the explanation. At training time, the gold label is provided, while at test time, we use the label predicted by the classifier. 
This architecture is depicted in \Cref{fig:model}.

\textbf{Loss   } We use negative log-likelihood for both classification and explanation losses. The explanation loss is much larger in magnitude than the classification loss, due to the summation of negative log-likelihoods over the words in the explanations. To account for this difference during training, we use a weighting coefficient $\alpha \in [0,1]$. Hence, our overall loss is: 
\begin{equation} 
  \mathcal{L}_\text{total} = \alpha \mathcal{L}_\text{label} + (1 - \alpha) \mathcal{L}_\text{explanation}
\end{equation}

\textbf{Selection   } We consider $\alpha$ values from $0.1$ to $0.9$ with a step of $0.1$ and decoder internal sizes of $512$, $1024$, $2048$, and $4096$. For this experiment, we choose as model selection criterion the accuracy on the SNLI validation set, because we want to investigate how well a model can generate justifications without sacrificing accuracy. As future work, one can inspect different trade-offs between accuracy and explanation generation. 
We found $\alpha=0.6$ and the decoder size of $512$ to produce the best validation accuracy, of $84.37\%$, while InferSent with no explanations produced $84.30\%$ validation accuracy. We call our model \eInferSent{}, since it freezes the InferSent architecture and training procedure, and only adds the explanations decoder.

\textbf{Results   }
The average test accuracy that we obtain when training InferSent\cite{infersent} on SNLI with five seeds is $84.01\%$  $(0.25)$. Our \eInferSent{} model obtains essentially the same test accuracy, of $83.96\%$ $(0.26)$, which shows that one can get additional justifications without sacrificing label accuracy. For the generated explanations, we obtain a perplexity of $10.58 (0.4)$ and a BLEU-score of $22.40 (0.7)$. Since we collected 3 explanations for each example in the validation and test sets, we compute the inter-annotator BLEU-score of the third explanation with respect to the first two, and obtain $22.51$. For consistency, we used the same two explanations as the only references when computing the BLEU-score for the predicted explanations. Given the low inter-annotator score and the fact that generated explanations almost match the inter-annotator BLEU-score, we conclude that this measure is not reliable for our task, and we further rely on human evaluation. Therefore, we manually annotated the first 100 datapoints in the test set (we used the same partial scoring as in Section \ref{partial-score}). Since the explanation is conditioned on the predicted label, for incorrect labels, the model would not produce a correct explanation. Therefore, we provide as correctness score the percentage of correct explanations in the subset of the first 100 examples where the predicted label was correct (80 in this experiment). We obtain a percentage of $34.68\%$ correct explanations. While this percentage is low, we keep in mind that the selection criteria was only the accuracy of the label classifier and not the perplexity of the explanation. In the next experiment, we show how training (and selecting) only for generating explanations results in higher quality explanations.

\subsection{\exptwo{}: Generate an explanation then predict a label}

In \expone, we conditioned the explanation on the label predicted by the MLP, because we wanted to see how the typical architecture used on SNLI can be adapted to justify its decisions in natural language. However, a more natural approach for solving inference is to think of the explanation first and based on the explanation to decide a label. Therefore, in this experiment, we first train a network to generate an explanation given a pair of (premise, hypothesis), and, separately, we train a network to provide a label given an explanation. This is a sensible decomposition for our dataset, due to the following key observation: In our dataset, in the large majority of the cases, one can easily detect for which label an explanation has been provided. We highlight that this is not the case in general, as the same explanation can be correctly arguing for different labels, depending on the premise and hypothesis. For example, the explanation \textit{"A woman is a person"} would be a correct explanation for the entailment pair \textit{("A woman is in the park", "A person is in the park")} as well for the contradiction pair \textit{("A woman is in the park", "There is no person in the park")}. However, there are multiple ways of formulating an explanation. In our example, for the contradiction pair, one could also explain that \textit{"There cannot be no person in the park if a woman is in the park"}, which read alone would allow one to infer that the pair was a contradiction. To support our observation, we train a neural network that given only an explanation predicts a label. We use the same bidirectional encoder and MLP-classifier as above. We obtain an accuracy of $96.83\%$ on the test set of SNLI.  

\textbf{Models   } For predicting an explanation given a pair of (premise, hypothesis), we first train a simple seq2seq model that we call \exptwoseqtoseq. Essentially, we keep the architecture in \eInferSent{}, where we eliminate the classifier by setting $\alpha=0$, and we decode the explanation without prepending the label. Secondly, we train an attention model, which we refer to as \exptwoattention. Attention mechanisms in neural networks brought consistent improvements over the non-attention counter-parts in various areas, such as computer vision \cite{showattend}, speech \cite{listenattend}, or natural language processing \cite{best-attention, BahdanauCB14}. We use the same encoder and decoder as in \exptwoseqtoseq, and we add two identical but separate attention modules, over the tokens in the premise and hypothesis. For details of the attention modules, see Appendix \ref{attention}.

\textbf{Selection   } Our only hyper-parameter is internal sizes for the decoder of $512$, $1024$, $2048$, and $4096$. Our model selection criterion is the perplexity on the validation set of SNLI. We obtain the best configuration for both \exptwoseqtoseq{} and \exptwoattention{} to have an internal size of $1024$.

\textbf{Results   } With the described setup, the SNLI test accuracy drops from $83.96\%$ $(0.26)$ in \expone{} to $81.59\%$ $(0.45)$ in \exptwoseqtoseq{} and $81.71\% (0.36)$ in \exptwoattention. 
However, when we again manually annotate the first 100 generated explanations in the test set, we obtain significantly higher percentages of correct explanations: $49.8\%$ for \exptwoseqtoseq{} and $64.27\%$ for \exptwoattention. We note that the attention mechanism indeed significantly increases the quality of the explanations. The perplexity and BLEU-score are $8.95 (0.03)$ and $24.14 (0.58)$ for \exptwoseqtoseq{}, and $6.1 (0)$ and $27.58 (0.47)$  for \exptwoattention. Our experiment shows that, while sacrificing a bit of performance, we get a better trust that when \exptwo{} predicts a correct label, it does so for the right reasons.

\textbf{Qualitative analysis of explanations   } In \Cref{generated-expls}, we provide examples of generated explanations from the test set from (a) \expone, (b) \exptwoseqtoseq{}, and (c) \exptwoattention. At the end of each explanation, we give in brackets the score that we manually allocated as explained in section \ref{partial-score}. We notice that the explanations are mainly on topic for all the three models, with minor exceptions, such as the mention of "\textit{camouflage}" in (1c). We also notice that even when incorrect, they are sometimes frustratingly close to being correct, for example, explanation (2b) is only one word (out of its 20 words) away from being correct. It is also interesting to inspect the explanations provided when the predicted label is incorrect. For example, in (1a), we see that the network omitted the information of "\textit{facing the camera}" in the premise and therefore classified the pair as neutral, which is backed up by an otherwise correct explanation in itself. We also see that model \exptwoseqtoseq{} correctly classifies this pair as entailment, however, it only motivates 1 out the 3 reasons why it is so, and it also picks arguably the easiest reason. Interestingly, the attention model (1c) points to the correct evidence but argues that "\textit{standing}" and "\textit{facing a camera}" is not enough to conclude "\textit{posing for a picture}". 


\begin{table}[]
\caption{Examples of predicted labels and generated explanations from (a)~\expone{},\\ (b)~\exptwoseqtoseq{}, and (c)~\exptwoattention. In brackets our scores of correctness. }

\smallskip 
\label{generated-expls}
\resizebox{\textwidth}{!}{%
\begin{tabular}{lll}
\toprule

\multicolumn{3}{c}{\begin{tabular}[c]{@{}c@{}} (1) Premise: 3 young man in hoods standing in the middle of a quiet street facing the camera. \\ Hypothesis: Three hood wearing people pose for a picture. \\ Gold label: entailment\end{tabular}} \\  

\\

\multicolumn{1}{l|}{\begin{tabular}[c]{@{}l@{}}(a) Predicted label: neutral \\ Explanation: Just because the men are in the \\ middle of a street doesn't mean \\ they are posing for a picture. [0] \end{tabular}}                                & \multicolumn{1}{l|}{\begin{tabular}[c]{@{}l@{}}(b) Predicted label: entailment \\ Explanation: three young men \\ are people. [0.33] \end{tabular}}                                                                       & \begin{tabular}[c]{@{}l@{}}(c) Predicted label: neutral \\ Explanation: Just because three young man \\ in camouflage standing in the middle of \\ a quiet street facing the camera \\ does not mean they pose for a picture. [0] \end{tabular}  \\

\\ \midrule

\multicolumn{3}{c}{\begin{tabular}[c]{@{}c@{}}(2) Premise: Three firefighter come out of subway station.\\ Hypothesis: Three firefighters putting out a fire inside of a subway station. \\ Gold label: neutral\end{tabular}} \\                             

\\

\multicolumn{1}{l|}{\begin{tabular}[c]{@{}l@{}}(a) Predicted label: contradiction \\ Explanation: The firefighters \\ can not be putting out a fire station and \\ putting out a fire at the same time. [0] \end{tabular}} & \multicolumn{1}{l|}{\begin{tabular}[c]{@{}l@{}}(b) Predicted label: neutral\\ Explanation: The fact that three\\ firemen are putting out of a subway\\ station doesn't imply that they\\  are putting out a fire. [0]\end{tabular}} & \begin{tabular}[c]{@{}l@{}}(c) Predicted label: neutral\\ Explanation: The firefighters \\ may not be putting out a fire \\ inside of the subway station. [1]\end{tabular} \\

\\  \midrule

\multicolumn{3}{c}{\begin{tabular}[c]{@{}c@{}} (3) Premise: A blond-haired doctor and her African American assistant looking threw new medical manuals.\\ Hypothesis: A man is eating pb and j.\\ Gold label: contradiction\end{tabular}}   \\                                                                                \\

\multicolumn{1}{l|}{\begin{tabular}[c]{@{}l@{}}(a) Predicted label: contradiction\\ Explanation: A man is not a woman. [1] \end{tabular}}                                                                                   & \multicolumn{1}{l|}{\begin{tabular}[c]{@{}l@{}}(b) Predicted label: contradiction\\ Explanation: One can not be looking \\ and eating simultaneously. [0] \end{tabular}}                                                               & \begin{tabular}[c]{@{}l@{}}(c) Predicted label: contradiction\\ Explanation: A person can not be looking \\ at a medical and a book at the same time. [0]
\end{tabular} 
\\
\bottomrule
\end{tabular}%
}
\end{table}

\subsection{\expthree{}: Universal sentence representations}

Obtaining universal sentence representations is the task of learning an encoder to provide semantically meaningful fixed-length representations of phrases/sentences. 
These representations are further used as features in other downstream tasks, particularly for tasks where labeled training data are scarce. 
In computer vision, pretrained ImageNet-based encoders \cite{imagenet} provide standard image feature extractors for other downstream tasks. 
However, in natural language processing, there is still no consensus on general-purpose sentence encoders. 
It remains an open question on which task and dataset should such an encoder be trained. Traditional approaches make use of very large unsupervised datasets \cite{skip, fastsent}, taking weeks to train. 
\citet{infersent} showed that training only on NLI is both more accurate and more time-efficient than training on orders of magnitude larger but unsupervised datasets. Their results constitute a previous state-of-the-art for universal sentence representations and encourage the idea that supervision can be more beneficial than larger but unsupervised datasets. We hypothesize that an additional layer of supervision in the form of natural language explanations should further improve learning of universal sentence representations.

\textbf{Model   } We use our \eInferSent{} model already trained in \expone. While we compare our model with InferSent that has not been trained on explanations, we want to ensure that eventual improvements are not purely due to the addition of a language model in the decoder network. 
We therefore introduce a second baseline, \inferSentAutoencoder, where instead of decoding explanations, we decode the premise and hypothesis separately from each sentence representation using one shared decoder.

\textbf{Evaluation metrics   }
Typically, sentence representations are evaluated by using them as fixed features on top of which shallow classifiers are trained for a series of downstream tasks. \citet{infersent} provide an excellent tool, called SentEval, for evaluating sentence representations on 10 diverse tasks: movie reviews ({\bf MR}), product reviews ({\bf CR}), subjectivity/objectivity ({\bf SUBJ}), opinion polarity ({\bf MPQA}), question-type ({\bf TREC}), sentiment analysis ({\bf SST}), semantic textual similarity ({\bf STS}), paraphrase detection ({\bf MRPC}), entailment ({\bf SICK-E}), and semantic relatedness ({\bf SICK-R}). 
We refer to their work for a more detailed description of each of these tasks and of SentEval, which we use for comparing the quality of the sentence embeddings obtained by additionally providing our explanations on top of the label supervision.

\textbf{Results   } In Table \ref{senteval}, we report the average results and standard deviations of \eInferSent, our retrained InferSent model, and the additional \inferSentAutoencoder{} baseline on the downstream tasks mentioned above. To test if the differences in performance of \inferSentAutoencoder{} and \eInferSent{} relative to the InferSent baseline are significant, we performed  Welch’s t-test.\footnote{Using the implementation in  \texttt{scipy.stats.ttest\_ind} with \texttt{equal\_var=False}.} We mark with * the results that appeared significant under the significance level of $0.05$.

\begin{table}
  \caption{Transfer results on downstream tasks. For MRPC we report accuracy/F1 score, for STS14 we report the Person/Spearman correlations, for SICK-R the Person correlation, and for all the rest their accuracies. Results are the average of 5 runs with different seeds. The standard deviations is shown in brackets, and the best result for every task is indicated in bold. * indicates significant difference at level 0.05 with respect to the InferSent baseline.}
  
  \smallskip 
  \label{senteval}
  \centering
  \begin{adjustbox}{width=1\textwidth}
  \begin{tabular}{lllllllllll}
    \toprule
    Model     & MR & CR & SUBJ & MPQA & SST2 & TREC & MRPC & SICK-E & SICK-R & STS14 \\
    \midrule
    InferSent-SNLI-ours & \textbf{78.18}  &	81.28 &	\textbf{92.46} &	88.46 &	\textbf{82.12} &	89.32 &	74.82 /	82.74 &	\textbf{85.96} &	0.887 &	0.65 / 0.63 \\
& (0.25) & (0.15) &	(0.15) &	(0.21) &	(0.22) &	(0.5) &	(0.66 /	0.27) &	(0.32) &	(0.002) &	(0 / 0) \\
    \inferSentAutoencoder &  75.94* &	79.26* &	91.72* &	88.16 &	80.9* &	\textbf{90.52*} &	\textbf{76.2*} /	82.48 &	85.58 &	0.88* &	0.5* / 0.5* \\
& (0.18) &	(0.36) &	(0.28) &	(0.26) &	(0.48) &	(0.52) &	(0.93 /	1.23) &	(0.33) &	(0) &	(0.02 / 0.02)\\
    \eInferSent & 77.76	&  \textbf{81.3} &	92.14* &	\textbf{88.78*} &	81.84 &	90 &	75.56 /	\textbf{83.24*} &	85.92 &	\textbf{0.89*} &	\textbf{0.68} / \textbf{0.65*} \\
& (0.44) &	(0.16) &	(0.21) &	(0.22) &	(0.4) &	(0.51) &	(0.62 / 0.24) &	(0.52) &	(0) &	(0.01 / 0.01)\\
    \bottomrule
  \end{tabular}
  \end{adjustbox}
\end{table}

We notice that \inferSentAutoencoder{} is performing significantly worse than InferSent on 6 tasks and significantly outperforms this baseline on only 2 tasks. This indicates that just adding a language generator can harm performance. Instead, \eInferSent{} significantly outperforms InferSent on 4 tasks, while it is significantly outperformed only on 1 task. Therefore, we conclude that training with explanations helps the model to learn overall better sentence representations.

\subsection{\expfour{}: Transfer without fine-tuning to out-of-domain NLI}
\label{exp4}

Transfer without fine-tuning to out-of-domain entailment datasets is known to exhibit poor performance. For example, \citet{snli} obtained an accuracy of only $46.7\%$ when training on SNLI and evaluating on SICK-E \cite{sick}. We test how our explanations affect the direct transfer in both label prediction and explanation generation by looking at SICK-E~\cite{sick} and MultiNLI~\cite{multinli}. The latter includes a diverse range of genres of written and spoken English, as well as test sets for cross-genre transfer.

\textbf{Model   } We again use our already trained \eInferSent{} model from \expone.

\textbf{Results   } In Table \ref{artifacts}, we present the performance of \eInferSent{} and our 2 baselines when evaluated without fine-tuning on SICK-E and MultiNLI. We notice that the accuracy improvements obtained with \eInferSent{} are very small. However, \eInferSent{} additionally provides explanations, which could bring insight into the inner workings of the model. We manually annotated the first 100 explanations of the test sets. The percentage of correct explanations in the subset where the label was predicted correctly was $30.64\%$ for SICK-E and only $1.92\%$ for MultiNLI.
We also noticed that the explanations in SICK-E, even when wrong, were generally on-topic and valid statements, while the ones in MultiNLI were generally nonsense or off-topic. 
Therefore, transfer learning for generating explanations in out-of-domain NLI would constitute challenging future work.

\begin{wraptable}{R}{0.5\textwidth}
  \caption{The average performance over 5 seeds of \eInferSent{} and the 2 baselines on SICK-E and MultiNLI with no fine-tuning. Standard deviations are in parenthesis. }
  \label{artifacts}
  \centering
  \resizebox{1\linewidth}{!}{
  \begin{tabular}{lll}
    \toprule
    Model     & SICK-E & MultiNLI \\
    \midrule
    InferSent-SNLI-ours & 53.27	(1.65) &	57 (0.41) \\
    
    \inferSentAutoencoder &  52.9 (1.77) & 55.38 (0.9) \\

    \eInferSent & \textbf{53.54} (1.43)	 & \textbf{57.16} (0.51)  \\
    
    \bottomrule
  \end{tabular}
  }
\end{wraptable}

\section{Related work}

\textbf{Interpretability   }
One main direction in interpretability for neural networks is providing extractive justifications, i.e., explanations consisting of subsets of the raw input, such as words or image patches. 
Extractive techniques can be divided into post-hoc (applied after training) and architecture-incorporated (guiding the training). 
For example, \citet{why-trust-you-RibeiroSG16} introduce a post-hoc extractive technique, LIME, that explains the prediction of any classifier via a local linear approximation around the prediction. 
\citet{tommi} introduces a similar approach but for structured prediction, where a variational autoencoder provides relevant perturbations of the inputs that are then used to infer pairs of input-output tokens that are causally related. 
While these models provide valuable insight for detecting biases, further model and dataset refinements would have to be made on a case-by-case basis. 
For example, \citet{artifacts} identified a set of biases in SNLI, but noted that their attempts to remove them would give rise to other biases. 

Attention-based models, such as \cite{BahdanauCB14,RocktaschelGHKB15}, offer some degree of interpretability and have been shown to also improve performance on downstream tasks. However, soft attention, the most prominent attention model, often does not learn to single out human-interpretable inputs. 

Neither extractive nor attention-based techniques can provide full-sentence explanations of a model's decisions.
Moreover, they cannot capture fine-grained relations and asymmetries, especially in a task like recognizing textual entailment. For example, if the words \textit{person, woman, mountain, outdoors} are extracted as justification, one may not know whether the model correctly learned that \textit{A woman is a person} and not that \textit{A person is a woman}, let alone that the model correctly paired (woman, person) and (mountain, outside).

\textbf{Natural language explanations   } 
In our work, we have taken a step further and built a neural network that is able to directly provide full-sentence natural language justifications. 
There has been little work on incorporating and outputting natural language free-form explanations, mostly due to the lack of appropriate datasets. 
In this direction, and very similar to our approach, is the recent work by \citet{zeynep}, who introduce two datasets of natural language explanations for the tasks of visual question-answering and activity recognition. 
Another work in this direction is that of \citet{math}, who introduced a dataset of textual justifications for solving math problems and formulate the task in terms of program execution. 
Nonetheless, their setup is specific to the task of solving math problems, and thus hard to transfer to more general natural understanding tasks. 
\citet{world-tree} provided a dataset of natural language explanation graphs for elementary science questions. 
However, with only $1,680$ pairs of questions and explanations, their corpus is orders of magnitude smaller than e-SNLI.

\textbf{Breaking natural language inference   }
Recently, an increasing amount of analysis has been carried out on the SNLI dataset and on the inner workings of different models trained on it. 
For example, \citet{dasgupta2018evaluating} assembled a dataset to test whether inference models actually capture compositionality beyond word level. 
They showed that InferSent sentence embeddings \cite{infersent}  indeed do not exhibit significant compositionality and that downstream models using these sentence representations largely rely on simple heuristics that are ecologically valid in the SNLI corpus. For example, high overlap in words between premise and hypothesis usually predicts entailment, while most contradictory sentence pairs have no or very little overlap of words. Negation words would also strongly indicate a contradiction.
\citet{breaking-nli} introduce a toy dataset, BreakingNLI, to test whether natural language inference models capture world knowledge and generalize beyond statistical regularities. 
To construct BreakingSNLI, they modified some of the original SNLI sentences such that they differ by at most one word from the sentences in the training set. 
\citeauthor{breaking-nli} show that models achieving high accuracies on SNLI, such as \cite{snli-1, snli-2, esim}, show dramatically reduced performance on this simpler dataset, while the model of \citet{kim} is more robust due to incorporating external knowledge. As the explanations in e-SNLI are mostly self-contained, our dataset provides the precise external knowledge that one requires in order to solve the SNLI inference task. It is therefore a perfect testbed for developing models that incorporate external knowledge from free-form natural language.

\section{Conclusions and future work}
We introduced e-SNLI, a large dataset of natural language explanations for an influential task of recognizing textual entailment. 
To demonstrate the usefulness of e-SNLI, we experimented with various ways of using these explanations for outputting human-interpretable full-sentence justifications of classification decisions. We also investigated the usefulness of these explanations as an additional training signal for learning better universal sentence representations and the transfer capabilities to out-of-domain NLI datasets.
In this work, we established a series of baselines using straight-forward recurrent neural network architectures for incorporating and generating natural language explanations. We hope that e-SNLI will be valuable for future research on more advanced models that would outperform our baselines.

Finally, we hope that the community will explore the dataset in other directions. For example, we also recorded the highlighted words, which we release with the dataset. Similar to the evaluation performed for visual question answering in \citet{humanVQA}, our highlighted words could provide a source of supervision and evaluation for attention models~\cite{RocktaschelGHKB15, snli-2} or post-hoc explanation models where the explanation consists of a subset of the input.



\bibliographystyle{apalike}  
\bibliography{bibfile}

\begin{appendices}

\section{List of templates to filter uninformative explanations} 
\label{expl_templates}

\textbf{General templates}

"<premise>" 

"<hypothesis>" 

"<hypothesis> <premise>" 

"<premise> <hypothesis>"

"Sentence 1 states <premise>. Sentence 2 is stating <hypothesis>"

"Sentence 2 states <hypothesis>. Sentence 1 is stating <premise>"

"There is <hypothesis>" 

"There is <premise>"

\textbf{Entailment templates} 

"<premise> implies <hypothesis>"

"If <premise> then <hypothesis>"

"<premise> would imply <hypothesis>"

"<hypothesis> is a rephrasing of <premise>"

"<premise> is a rephrasing of <hypothesis>" 

"In both sentences <hypothesis>"

"<premise> would be <hypothesis>"

"<premise> can also be said as <hypothesis>"

"<hypothesis> can also be said as <premise>"

"<hypothesis> is a less specific rephrasing of <premise>"

"This clarifies that <hypothesis>"

"If <premise> it means <hypothesis>"

"<hypothesis> in both sentences"

"<hypothesis> in both"

"<hypothesis> is same as <premise>"

"<premise> is same as <hypothesis>"

"<premise> is a synonym of <hypothesis>"

"<hypothesis> is a synonym of <premise>".

\textbf{Neutral templates} 

"Just because <premise> doesn't mean <hypothesis>"

"Cannot infer the <hypothesis>"

"One cannot assume <hypothesis>"

"One cannot infer that <hypothesis>"

"Cannot assume <hypothesis>"

"<premise> does not mean <hypothesis>"

"We don't know that <hypothesis>"

"The fact that <premise> doesn't mean <hypothesis>"

"The fact that <premise> does not imply <hypothesis>"

"The fact that <premise> does not always mean <hypothesis>"

"The fact that <premise> doesn't always imply<hypothesis>".

\textbf{Contradiction templates}

"In sentence 1 <premise> while in sentence 2 <hypothesis>"

"It can either be <premise> or <hypothesis>"

"It cannot be <hypothesis> if <premise>"

"Either <premise> or <hypothesis>" 

"Either <hypothesis> or <premise>" 

"<premise> and other <hypothesis>"

"<hypothesis> and other <premise>"

"<hypothesis> after <premise>"

"<premise> is not the same as <hypothesis>"

"<hypothesis> is not the same as <premise>" 

"<premise> is contradictory to <hypothesis>" 

"<hypothesis> is contradictory to <premise>" 

"<premise> contradicts <hypothesis>"

"<hypothesis> contradicts <premise>"

"<premise> cannot also be <hypothesis>"

"<hypothesis> cannot also be <premise>"

"either <premise> or <hypothesis>" 

"either <premise> or <hypothesis> not both at the same time"

"<premise> or <hypothesis> not both at the same time".

\section{Architecture of \exptwoattention} 
\label{attention}
Our attention model \exptwoattention{} is composed of two identical but separate modules for premise and hypothesis. We fix the number of attended tokens at $84$, the maximum length of a sentence in SNLI. We denote by $h_t^p$ and $h_t^h$ the bidirectional embeddings of the premise and hypothesis at timestep $t$. We denote by $h_{\tau}^{dec}$ the decoder hidden state at timestep $\tau$, which we refer to as the context of the attention.

We use 3 couples of linear projections followed by $tanh$ non-linearities as follows:

We project each timestep of the encoder for premise and hypothesis:
$$ proj1_t^p = tanh(W^1_p h_t^p + b^1_p) $$ $$proj1_t^h = tanh(W^1_h h_t^h + b^1_h). $$

We separately project the context vector, that is, the hidden vector of the decoder at each timestep, before doing its dot product with the tokens of the premise and hypothesis:
$$proj_{\tau}^{c, p} = tanh(W^c_p h_{\tau}^{dec} + b^c_p) $$ 
$$ proj_{\tau}^{c, h} = tanh(W^c_h h_{\tau}^{dec} + b^c_h).$$

At each decoding timestep $\tau$, we do the dot product between the projections of the context with all the timesteps of the premise and hypothesis, respectively:
$$ \widetilde{w_t}^{p, \tau} = <proj_{\tau}^{c, p} , proj1_t^p> $$ 
$$ \widetilde{w_t}^{h, \tau} = <proj_{\tau}^{c, h} , proj1_t^h>. $$

The final attention weights are computed from a softmax over the non-normalized weights:
$$ w_t^{p, \tau} = Softmax(\widetilde{w_t}^{p, \tau}) $$ 
$$w_t^{h, \tau} = Softmax(\widetilde{w_t}^{h, \tau}).  $$

We use another couple of projections for the embeddings of the tokens of premise and hypothesis, before we apply the weighted sum. 
$$ proj2_t^p = tanh(W^2_p h_t^p + b^2_p) $$ $$ proj2_t^h = tanh(W^2_h h_t^h + b^2_h). $$

Finally, we compute the weighted sums for premise and hypothesis:

$$ p_{\tau} = \sum_t w_t^{p, \tau} proj2_t^p$$ 
$$h_{\tau} = \sum_t w_t^{h, \tau} proj2_t^h.
 $$

At each timestept $\tau$, we concatenate $p_{\tau}$ and $h_{\tau}$ with the word embedding from the previous timestep $\tau - 1$ and give as input to our decoder.

\end{appendices}

\end{document}